\documentclass[runningheads]{llncs}
\usepackage{graphicx}
\usepackage{color}
\usepackage{amsmath}
\usepackage{amssymb}
\usepackage{amsfonts}
\usepackage{xspace}
\usepackage{wrapfig}
\usepackage{tikz}
\usepackage{environ}
\usepackage{subcaption}
\usepackage{algorithm}
\usepackage{algpseudocode}
\usepackage{hyperref}    
\usepackage{url}     
\usepackage{cite}
\newcommand{\R}{\mathbb{R}}
\newcommand{\seg}{\mathbf{y}}
\newcommand{\y}{\mathbf{y}}

\newcommand{\x}{\mathbf{x}}
\newcommand{\X}{\mathbf{X}}
\newcommand{\w}{\mathbf{w}}
\newcommand{\A}{\mathbf{A}}

\begin{document}
\title{Is segmentation uncertainty useful?}
%
%
\author{%
Steffen Czolbe\inst{\dagger1}  \and 
Kasra Arnavaz\inst{\dagger1} \and 
Oswin Krause\inst{1} \and 
Aasa Feragen\inst{2}
}%
\institute{
University of Copenhagen, Department of Computer Science, Denmark,\\
\texttt{per.sc,kasra,oswin.krause@di.ku.dk}
\and
Technical University of Denmark, DTU Compute, Denmark,\\
\texttt{afhar@dtu.dk} \\
$\dagger$ Authors contributed equally\\
}
\maketitle              
\begin{abstract}
Probabilistic image segmentation encodes varying prediction confidence and inherent ambiguity in the segmentation problem. While different probabilistic segmentation models are designed to capture different aspects of segmentation uncertainty and ambiguity, these modelling differences are rarely discussed in the context of applications of uncertainty.
We consider two common use cases of segmentation uncertainty, namely assessment of segmentation quality and active learning. We consider four established strategies for probabilistic segmentation, discuss their modelling capabilities, and investigate their performance in these two tasks. We find that for all models and both tasks, returned uncertainty correlates positively with segmentation error, but does not prove to be useful for active learning.

\end{abstract}

\section{Introduction}

\makeatletter{\renewcommand*{\@makefnmark}{} 
\footnotetext{Code available at \href{https://github.com/SteffenCzolbe/probabilistic\_segmentation}{\texttt{github.com/SteffenCzolbe/probabilistic\_segmentation}}}\makeatother}

Image segmentation -- the task of delineating objects in images -- is one of the most crucial tasks in image analysis. As image acquisition methods can introduce noise, and experts disagree on ground truth segmentations in ambiguous cases, predicting a single segmentation mask can give a false impression of certainty. Uncertainty estimates inferred from the segmentation model can give some insight into the confidence of any particular segmentation mask, and highlight areas of likely segmentation error to the practitioner. It adds transparency to the segmentation algorithm and communicates this uncertainty to the user. This is particularly important in medical imaging, where segmentation is often used to understand and treat disease. Consequently, quantification of segmentation uncertainty has become a popular topic in biomedical imaging~\cite{kohl2018probabilistic,monteiro2020stochastic}. 

Training segmentation networks requires large amounts of annotated data, which are costly and cumbersome to attain. Active learning aims to save the annotator's time by employing an optimal data gathering strategy. Some active learning methods use uncertainty estimates to select the next sample to annotate~\cite{mackay1992c,lewis1994heterogeneous,mackay1992evidence}. While several potential such data gathering strategies exist~\cite{settles2009active,yang2018benchmark}, a consistent solution remains to be found~\cite{loog2016empirical}. 

While several methods have been proposed to quantify segmentation uncertainty~\cite{gal2016dropout,kohl2018probabilistic,monteiro2020stochastic}, it is rarely discussed what this uncertainty represents, whether it matches the user's interpretation, and if it can be used to formulate a data-gathering strategy.  We compare the performance of several well-known probabilistic segmentation algorithms, assessing the quality and use cases of their uncertainty estimates. We consider two segmentation scenarios: An unambiguous one, where annotators agree on one underlying true segmentation, and an ambiguous one, where a set of annotators provide potentially strongly different segmentation maps, introducing variability in the ground truth annotation.

We investigate the degree to which the inferred uncertainty correlates with segmentation error, as this is how reported segmentation uncertainty would typically be interpreted by practitioners. We find that uncertainty estimates of the models coincide with likely segmentation errors and strongly correlate with the uncertainty of a set of expert annotators. Surprisingly, the model architecture used does not have a strong influence on the quality of estimates, with even a deterministic U-Net \cite{ronneberger2015u} giving good pixel-level uncertainty estimates.

Second, we study the potential for uncertainty estimates to be used for selecting samples for annotation in active learning. Reducing the cost of data annotation is of utmost importance in biomedical imaging, where data availability is fast-growing, while annotation availability is not. We find that there are many pitfalls to an uncertainty-based data selection strategy. In our experiment with multiple annotators, the images with the highest model uncertainty were precisely those images where the annotators were also uncertain. Labeling these ambiguous images by a group of expert annotators yielded conflicting ground truth annotations, providing little certain evidence for the model to learn from.

\begin{figure}[t]
\centering
\includegraphics[width=\textwidth]{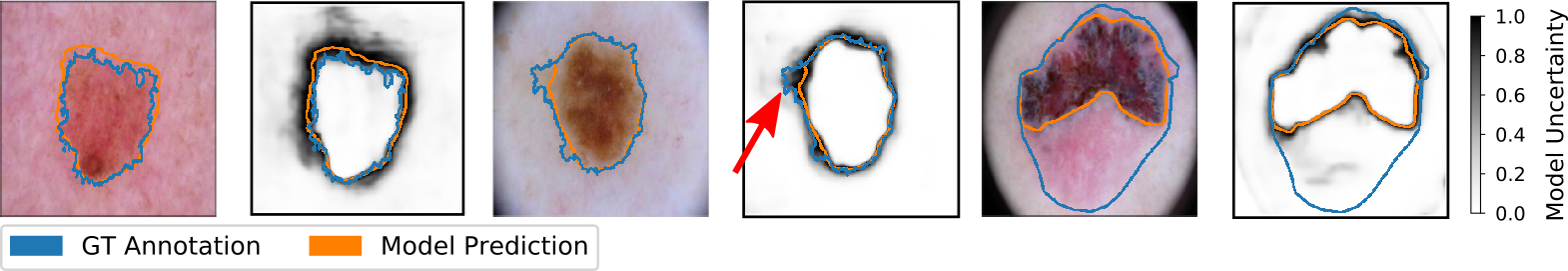}
\vspace{-1.5em}
\caption{Segmentation uncertainty is often interpreted as probable segmentation error, as seen near the lesion boundary in the first two examples. In the third example, however, model bias leads to a very certain, yet incorrect segmentation.}
\label{fig:init}
\end{figure}

\section{Modelling segmentation uncertainty}

Image segmentation seeks to estimate a well-defined binary\footnote{For simplicity, we consider binary segmentation; the generalization to multi-class segmentation is straightforward.} segmentation $g \colon \Omega \to \{0,1\}$ for a discrete image domain $\Omega$. Typically, a predictive model  $h(\x, \w)$ with parameters $\w$, such as a neural network, is fitted to binary annotation data $a \colon \Omega \to \{0,1\}$ by minimizing a loss $\mathcal{L}(a, h(\x, \w))$. Here, $\x \in \R^\Omega$ is the image, and $\y = h(\x, \w)$ defines an image of pixel-wise segmentation probabilities, such as the un-thresholded softmax output of a segmentation network $h$. 

Typically, the annotation is assumed to be error-free, that is $a = g$, and predictors are typically trained on a single annotation per image. We assume that the trained neural network $h(\x, \w)$ satisfies
\[
h(\x, \w) = g(\x) + b + err \enspace,
\]
where $b$ and $err$ denote bias and segmentation error. Segmentation uncertainty is often interpreted as correlating with this error, although this is primarily realistic for small bias. Such segmentation tasks are called \emph{unambiguous}; we consider a running example of skin lesion segmentation from dermoscopic images~\cite{isic1,isic2}, where the lesion boundary is clearly visible in the image (Fig.~\ref{fig:init}).

Recent work has considered \emph{ambiguous} segmentation tasks~\cite{kohl2018probabilistic,monteiro2020stochastic}, where there is no accessible ``ground truth'' segmentation, either because the data is not sufficient to estimate the segmentation, or because there is subjective disagreement. Examples include lesions in medical imaging, where the boundary can be fuzzy due to gradual infiltration of tissue, or where experts disagree on whether a tissue region is abnormal or not. 

In such tasks, we make no assumption on the underlying segmentation $g$ or the errors $err$, but regard the observed annotations as samples from an unknown ``ground truth" distribution $p(a|\x)$ over annotations $a$ conditioned on the image $\x$. The goal of segmentation is to estimate the distribution $p(a|\x)$, or its proxy distribution $p(\y|\x)$ over pixel-wise class probabilities $\y \colon \Omega \to [0,1]$, as accurately as possible for a given image $\x$. If successful, such a model can sample coherent, realistic segmentations from the distribution, and estimate their variance and significance. As a running example of an ambiguous segmentation task, we consider lung lesions~\cite{lidc1,lidc2,kohl2018probabilistic}. For such tasks, predictors are typically trained on multiple annotators, who may disagree both on the segmentation boundary and on whether there is even an object to segment. 

From the uncertainty modelling viewpoint, these two segmentation scenarios are rather different. Below, we discuss differences in uncertainty modelling for the two scenarios and four well-known uncertainty quantification methods.

\section{Probabilistic Segmentation Networks} \label{sec:networks}
A probabilistic segmentation model seeks to model the distribution $p(\y | \x)$ over segmentations given an input image $\x$. 
Here, our annotated dataset $(\X, \A)$ consists of the set $\X$ of $N$ images $\big\{\x_n~|~n=1,...,N\big\}$, and $L$ annotations are available per image, so that $\A=\big\{a_{n}^{(l)} \sim p(\seg |\x_n)~|~(n,l)=(1,1),...,(N,L)\big\}$. 

Taking a Bayesian view, we seek the distribution
\begin{equation} \label{integral}
        p(\mathbf y|\mathbf x, \mathbf X, \A) = \int p(\y| \mathbf x, \mathbf w) p(\mathbf w|\mathbf X, \A, h) \,d\mathbf w \enspace,
\end{equation}
over segmentations $\y$ given image $\x$ and data $(\X, \A)$, which can be obtained by marginalization with respect to the weights $\w$ of the model $h$. 

In most deep learning applications, our prior belief over the model $h$, denoted $p(h)$, is modelled by a Dirac delta distribution indicating a single architecture with no uncertainty. In the context of uncertain segmentation models, however, we would like to model uncertainty in the parameters $\w$. Denoting our prior belief over the parameters $\w$ by $p(\w|h)$, Bayes' theorem gives
\begin{equation} \label{post}
    p(\mathbf w|\mathbf X, \A, h) = \frac{p(\mathbf w|h)p(\A|\mathbf X, \mathbf w, h)}{p(\A|\mathbf X, h)},
\end{equation}
where the likelihood update function is given by
\begin{equation} \nonumber
    p(\A|\mathbf X, \mathbf w, h) = \exp{\left(\sum_{n=1}^{N} \sum_{l=1}^{L} \A^{(l)}\log \left( h(\mathbf x_n, \mathbf w)\right) + (1-\A^{(l)}) \log \left( 1-h(\mathbf x_n, \mathbf w)\right) \right)} 
\end{equation}
and normalizing constant
\begin{equation} \nonumber
p(\A|\mathbf X, h) = \int p(\mathbf w|h)p(\A|\mathbf X, \mathbf w, h)\,d\mathbf w \enspace .
\end{equation}

This integral is generally intractable, making it impossible to obtain the proper posterior~\eqref{post}. Below, we discuss how empirical approximations $\hat{p}(\mathbf y|\mathbf x, \mathbf X, \A)$ to the distribution $p(\mathbf y|\mathbf x, \mathbf X, \A)$ found in~\eqref{integral} are performed in four common segmentation models. Note that both $p$ and $\hat p$ can be degenerate, depending on the number of annotations available and models used.

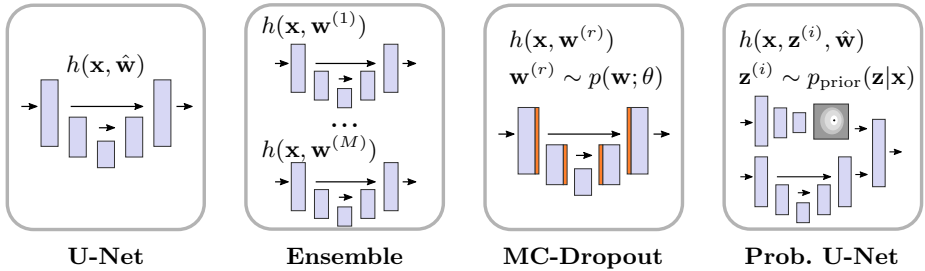
\begin{figure}[t]
	\centering
	\begin{tikzpicture}[scale=\textwidth/1cm,samples=200] 
	\begin{scope}
	\node[anchor=south west, inner sep=0pt] (image0) at (0,0) {\includegraphics[width=1.0\textwidth]{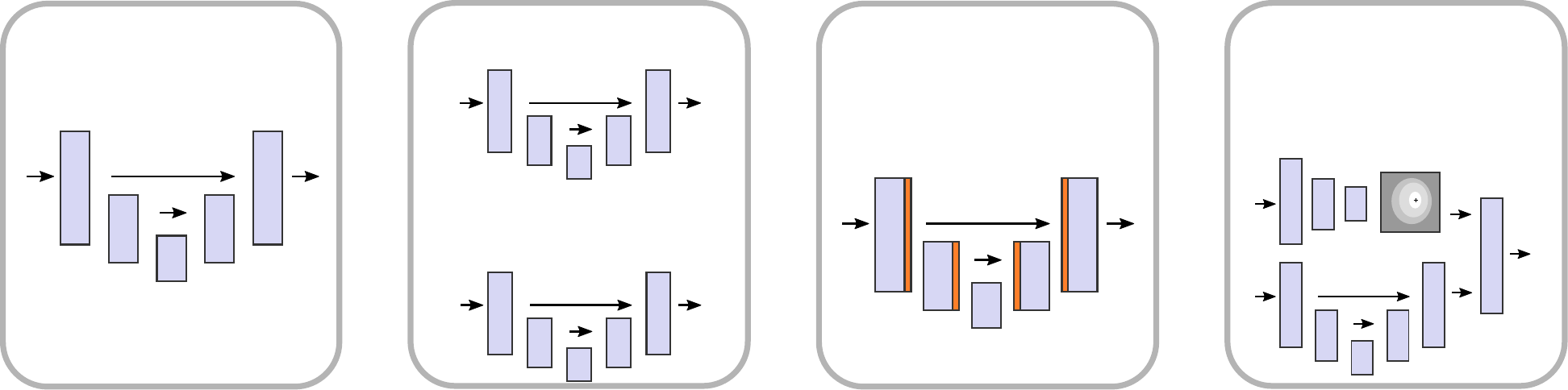}};
	\end{scope}
	
	\node [anchor=mid] (note) at (0.11,-0.03) {\textbf{U-Net}};
	\node [anchor=mid] (note) at (0.11,0.18) {$h(\x, \hat\w)$};
	
	\node [anchor=mid] (note) at (0.37,-0.03) {\textbf{Ensemble}};
	\node [anchor=west] (note) at (0.27,0.225) {$h(\x, \w^{(1)})$};
	\node [anchor=mid] (note) at (0.37,0.12) {\Large ...};
	\node [anchor=west] (note) at (0.27,0.09) {$h(\x, \w^{(M)})$};
	
	\node [anchor=west] (note) at (0.54,0.21) {$h (\mathbf x, \mathbf w^{(r)})$};
	\node [anchor=west] (note) at (0.54,0.17) {$\w^{(r)} \sim p(\w;\theta)$};
	\node [anchor=mid] (note) at (0.63,-0.03) {\textbf{MC-Dropout}};

	\node [anchor=west] (note) at (0.79,0.21) {$h(\x, \mathbf{z}^{(i)}, \hat \w)$};
	\node [anchor=west] (note) at (0.79,0.17) {$\mathbf{z}^{(i)} \sim p_{\text{prior}}(\mathbf{z} | \x)$};
	\node [anchor=mid] (note) at (0.89,-0.03) {\textbf{Prob. U-Net}};

	\end{tikzpicture}
	\vspace{-1.5em}
	\caption{Schematic overview (adapted from~\cite{kohl2018probabilistic}) of the evaluated models. Blue: residual blocks 
	. Orange: Dropout layers 
	essential to the networks' functionality.}
	\label{fig:model_architectures}
\end{figure}

\subsubsection*{U-Net with softmax output.}
The well established U-Net~\cite{ronneberger2015u} architecture with a softmax output layer yields class-likelihood estimates. As the model is deterministic, $p(h|\X, \A)$ is degenerate. Parameters are selected by a maximum a posteriori (MAP) estimate i.e. $p(\w |\X, \A, h) \approx \delta(\w-\hat{\w})$ in which $\hat \w = \text{argmax }p(\w|\X, \A, h)$. The model output \eqref{integral} is approximated by the degenerate distribution $\hat{p}(\y|\x, \X, \A) \approx  p(\y| \mathbf x, \hat \w)$.
The softmax output layer predicts a pixel-wise class probability distribution $p(\y_{(i,j)} | \x, \X, \A)$. As no co-variance or dependencies between pixel-wise estimates are available, segmentation masks sampled from the pixel-wise probability distributions are often noisy~\cite{monteiro2020stochastic}. An alternative approach followed by our implementation is the thresholding of pixel-wise probability values, which leads to a single, coherent segmentation map.

\subsubsection*{Ensemble methods} combine multiple models to obtain better predictive performance than that obtained by the constituent models alone, while also allowing the sampling of distinct segmentation maps from the ensemble. We combine $M$ U-Net models $h(\x, \w^{(m)})$ where, if labels from multiple annotators are available, each constituent model is trained on a disjoint label set $\A^{(m)}$. When trained on datasets with a single label, all constituent models are trained on the same data and their differences stem from randomized initialization and training. Treating the models as samples, we obtain an empirical distribution approximating~\eqref{integral} by drawing from the constituent models at random.

\subsubsection*{Monte-Carlo Dropout} \cite{gal2016dropout} is a Bayesian approximation technique based on dropout, where samples from the posterior over dropout weights give a better approximation of the true posterior than a MAP estimation. Given a selected model $h$, one can approximate \eqref{integral} as $
\hat p(\y|\x, \X, \A) \approx 1/R \sum_{r=1}^{R} p(\y| \mathbf x, \w^{(r)})$ when $\w^{(r)} \sim p(\mathbf w|\mathbf X, \A, h)$. Since $p(\mathbf w|\mathbf X, \A, h)$  is intractable, it is approximated~\cite{gal2016dropout} by a variational distribution $p(\theta)$ as $\theta_i = w_i \cdot z_i, z_i \sim \text{Bernoulli}(p_i)$, where $p_i$ is the probability of keeping the weight $w_i$ in a standard dropout scheme.

\subsubsection*{The Probabilistic U-Net} \cite{kohl2018probabilistic} fuses the output of a deterministic U-Net with latent samples from a conditional variational auto-encoder modelling the variation over multiple annotators. Test-time segmentations are formed by sampling a latent $\mathbf{z}$, which is propagated with the image through the U-Net. Predictions are made as $
    \hat p(\y|\x, \X, \A) \approx p(\y| \mathbf x, \mathbf z^{(i)}, \hat \w)$, with $\mathbf{z}^{(i)} \sim p_{\text{prior}}(\mathbf{z} | \x)$.

\begin{figure}[t]
	\centering
	\begin{tikzpicture}
	\node [anchor=mid, inner sep=0pt] (note) at (2.45,6.2) {Skin Lesions};
	\draw (0.03,6.05) -- (4.83,6.05);
	\node [anchor=mid, inner sep=0pt] (note) at (7.4,6.2) {Lung Cancer};
	\draw (4.95,6.05) -- (9.75,6.05);
	\node [anchor=mid, inner sep=0pt] (note) at (-1.3,5.25) {Image};
	\node [anchor=mid, inner sep=0pt] (note) at (-1.3,4.075) {\textbf{U-Net}};
	\node [anchor=mid, inner sep=0pt] (note) at (-1.3,2.9) {\textbf{Ensemble}};
	\node [anchor=mid, inner sep=0pt] (note) at (-1.3,1.725) {\textbf{MC-Dropout}};
	\node [anchor=mid, inner sep=0pt] (note) at (-1.3,0.55) {\textbf{Prob. U-Net}};
	\begin{scope}
	\node[anchor=south west, inner sep=0pt] (image0) at (0,0) {\includegraphics[width=0.8\textwidth]{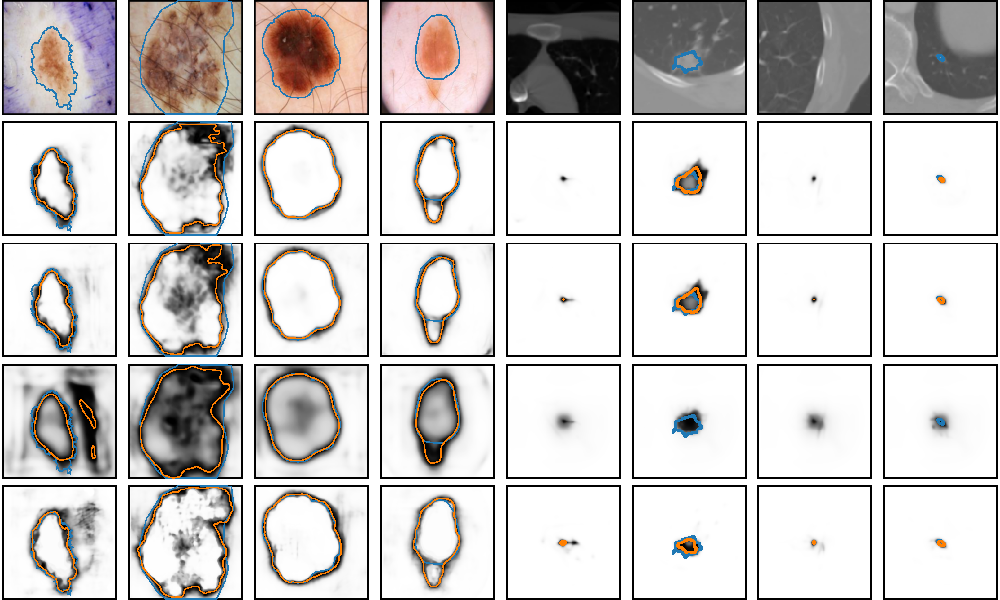}};
	\end{scope}
	
	\begin{scope}
	\node[anchor=south west, inner sep=0pt] (image1) at (0,-1.0) {\includegraphics[width=0.8\textwidth]{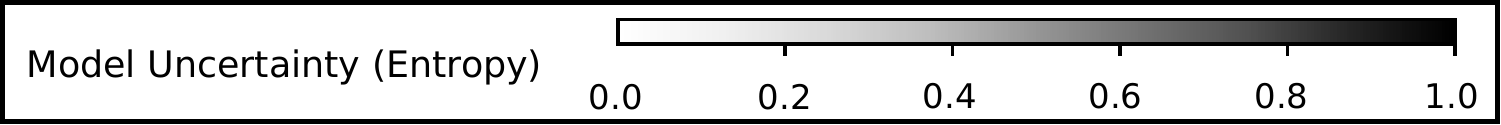}};
	\end{scope}
	
	\end{tikzpicture}
	\caption{Segmentation Uncertainty. Samples from the test set of the two datasets. Images in row one, model uncertainty (entropy) heat-maps in rows 2-5. Outline of mean ground truth annotations in Blue, mean model predictions in Orange.}
	\label{fig:uncertainty_samples}
\vspace{-1em}
\end{figure}
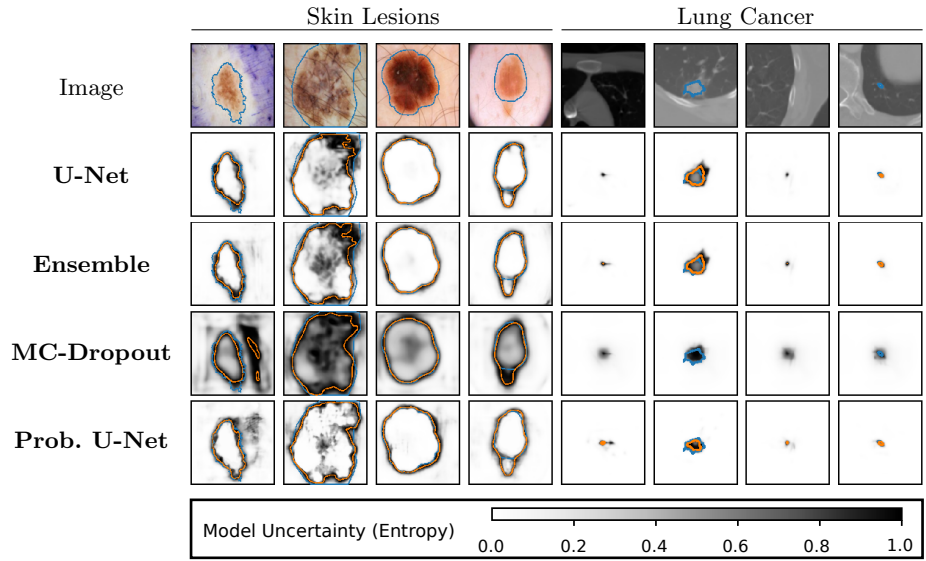

\section{Experiments}

\subsection{Data}
Practical applications of uncertainty in segmentation tasks differ both in the type of ambiguity, and the availability of expert annotations. We select two representative datasets for our evaluation.

The \textbf{ISIC18} dataset consists of skin lesion images with a single annotation available~\cite{isic1,isic2}, and is used as an example of unambiguous image segmentation. We rescale the images to $256 \times 256$ pixels and split the dataset into $1500$ samples for the train-set and $547$ each for the validation and test sets.

The \textbf{LIDC-IDRI} lung cancer dataset~\cite{lidc1,lidc2} contains 1018 lung CT scans from 1010 patients. For each scan, 4 radiologists (out of 12) annotated abnormal lesions. Anonymized annotations were shown to the other annotators, who were allowed to adjust their own masks. Significant disagreement remains between the annotators: Among the extracted patches where at least one annotator marked a lesion, an average of $50\%$ of the annotations are blank. 
We pre-processed the images as in~\cite{kohl2018probabilistic}, resampled to $0.5$mm resolution, and cropped the CT-slices with lesions present to $128 \times 128$ pixels. The dataset is split patient-wise into three groups, $722$ for the training-set and $144$ each for the validation and test sets.

\subsection{Model tuning and training}
To allow for a fair evaluation, we use the same U-Net backbone of four encoder and decoder blocks for all models. Each block contains an up-/down-sampling layer, three convolution layers, and a residual skip-connection. The ensemble consists of four identical U-Nets. The latent-space encoders of the probabilistic U-Net are similar to the encoding branch of the U-Nets, and we choose a six-dimensional latent space size, following the original paper's recommendation.

All models were trained with binary cross-entropy. The probabilistic U-Net has an additional $\beta$-weighted KL-divergence loss to align the prior and posterior distributions, as per \cite{kohl2018probabilistic}. The optimization algorithm was Adam, with a learning rate of $10^{-4}$ for most models, except the probabilistic U-Net and MC-Dropout models on the skin lesion dataset, where a lower learning rate of $10^{-5}$ gave better results. We utilized early stopping to prevent over-fitting, and define the stopping criteria as 10 epochs without improvement of the validation loss, 100 epochs for models trained with the reduced learning rate. For the MC-Dropout and probabilistic U-Net models we performed a hyper-parameter search over the dropout probability $p$ and the loss function weighting factor $\beta$, selecting the configuration with the lowest generalized energy distance on the validation set. We arrived at $p=0.5$, $\beta = 0.0005$.

\begin{figure}[t]
     \centering
     \begin{subfigure}[b]{0.49\textwidth}
         \centering
         \includegraphics[width=\textwidth]{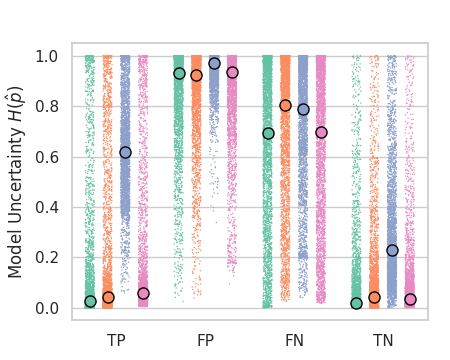}
         \caption{Skin Lesion Dataset}
     \end{subfigure}
     \hfill
     \begin{subfigure}[b]{0.49\textwidth}
         \centering
         \includegraphics[width=\textwidth]{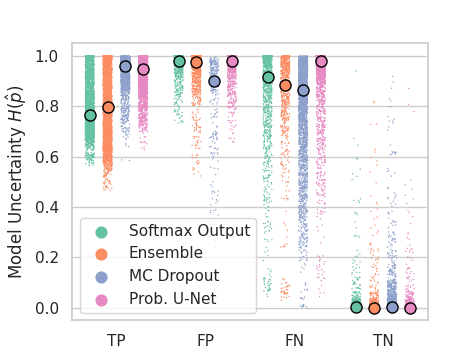}
         \caption{Lung Cancer Dataset}
     \end{subfigure}
     \hfill
    \caption{Pixelwise uncertainty by prediction correctness (True Positive, False Positive, False Negative, True Negative). The scatter plot shows individual pixels, with the median circled. For the lung cancer dataset, we discarded pixels with annotator disagreement.}
    \label{fig:uncertainty_by_condition}
\end{figure}

\begin{figure}
\centering
\begin{minipage}[t]{.48\textwidth}
  \includegraphics[width=1\textwidth]{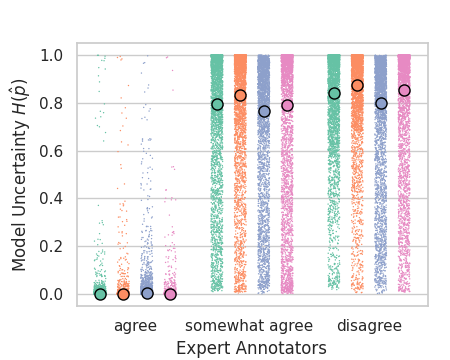}
  \captionof{figure}{Pixel-wise model uncertainty on the lung cancer dataset, grouped by agreement of expert annotations. Experts agree: $H(p) = 0$, somewhat agree $0 < H(p) < 1$, disagree $H(p) = 1$.}
  \label{fig:lidc_uncertainty_correl}
\end{minipage}%
\hfill
\begin{minipage}[t]{.48\textwidth}
  \includegraphics[width=1\textwidth]{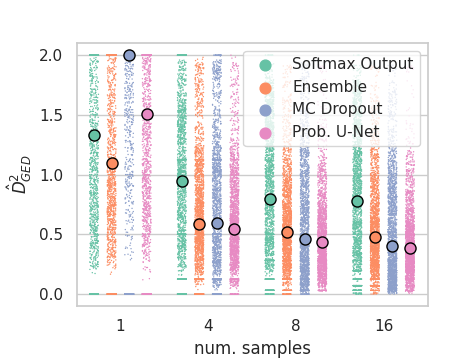}
  \captionof{figure}{Generalized Energy Distance of models on the lung cancer dataset, approximation by 1 to 16 samples, median highlighted. Lower distances are better.}
  \label{fig:lidc_ged}
\end{minipage}
\end{figure}

\subsection{Uncertainty Estimation}

For all models, our uncertainty estimates are based on non-thresholded  pixel-wise predictions. For the U-Net, we take the final softmax predictions; for the remaining models we average across $16$ non-thresholded samples. We quantify the \textit{pixel-wise} uncertainty of the model by the entropy
\begin{equation} \nonumber
\mathrm{H}(p(\seg_{(i,j)} | \x, \X, \A)) = \sum_{c \in C} p(\seg_{(i,j)} = c | \x, \X, \A) \log_2 \frac{1}{p(\seg_{(i,j)} = c | \x, \X, \A)}
\end{equation}
with $p(\seg_{(i,j)} = c | \x)$ as the pixel-wise probability to predict class $c \in C$. We plot the resulting uncertainty map for random images $\x$ from both datasets in Fig.~\ref{fig:uncertainty_samples}. For visual reference, we overlay the mean expert annotation in Blue, and the mean model prediction in Orange. Darker shades indicate higher uncertainty.

We quantitatively assess the quality of uncertainty estimates by examining their relation to segmentation error in Fig.~\ref{fig:uncertainty_by_condition}. On both datasets, models are more certain when they are correct (true positive, true negative) compared to when they are incorrect (false positive, false negative). A repeated measure correlation test finds a significant $(\alpha = 0.01)$ correlation between segmentation error and model uncertainty on both datasets, for all methods. The relation holds, but is less strong, for MC-dropout on the skin dataset, which retains high uncertainty even when it is correct. On the lung cancer dataset, all models have high uncertainty on true positive predictions. This might be caused by the imbalance of the dataset, where the positive class is strongly outweighed by the background and annotators often disagree. We tried training the models with a class-occurrence weighted loss function, which did produce true positive predictions with higher certainty but suffered an overall higher segmentation error.

We assess the correlation of model uncertainty with the uncertainty of the annotators on the lung cancer dataset in Fig.~\ref{fig:lidc_uncertainty_correl}. For all models, this correlation is significant $(\alpha = 0.01)$ . The median model uncertainty is very low ($< 0.1$) when all annotators agree, but high ($> 0.7$) when they disagree. There is a minor difference in model uncertainty between partial agreement (annotators split 3 -- 1) and full disagreement (annotators split 2 -- 2).

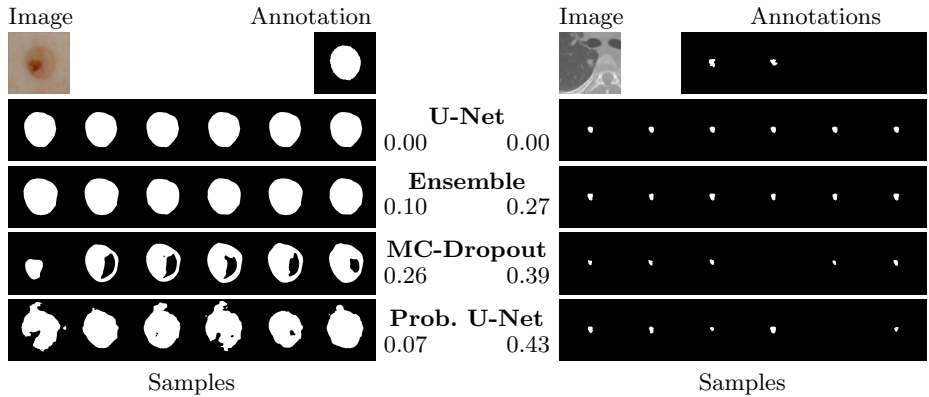
\begin{figure}[t]
	\centering
	\begin{tikzpicture}[scale=\textwidth/1cm,samples=200] 
	\node [anchor=west, inner sep=0pt] (note) at (0.0,0.375) {Image};
	\node [anchor=mid] (note) at (0.33,0.375) {Annotation};
	\node [anchor=west, inner sep=0pt] (note) at (0.6,0.375) {Image};
	\node [anchor=mid] (note) at (0.88,0.375) {Annotations};
	\node [anchor=mid] (note) at (0.2,-0.025) {Samples};
	\node [anchor=mid] (note) at (0.8,-0.025) {Samples};
	\node [anchor=mid] (note) at (0.5,0.265) {\textbf{U-Net}};
	\node [anchor=west] (note) at (0.4,0.24) {$0.00$};
	\node [anchor=east] (note) at (0.6,0.24) {$0.00$};
	\node [anchor=mid] (water) at (0.5,0.195) {\textbf{Ensemble}};
	\node [anchor=west] (note) at (0.4,0.17) {$0.10$};
	\node [anchor=east] (note) at (0.6,0.17) {$0.27$};
	\node [anchor=mid] (water) at (0.5,0.12) {\textbf{MC-Dropout}};
	\node [anchor=west] (note) at (0.4,0.095) {$0.26$};
	\node [anchor=east] (note) at (0.6,0.095) {$0.39$};
	\node [anchor=mid] (water) at (0.5,0.045) {\textbf{Prob. U-Net}};
	\node [anchor=west] (note) at (0.4,0.02) {$0.07$};
	\node [anchor=east] (note) at (0.6,0.02) {$0.43$};
	
	\begin{scope}
	\node[anchor=south west, inner sep=0pt] (image0) at (0,0) {\includegraphics[width=0.4\textwidth]{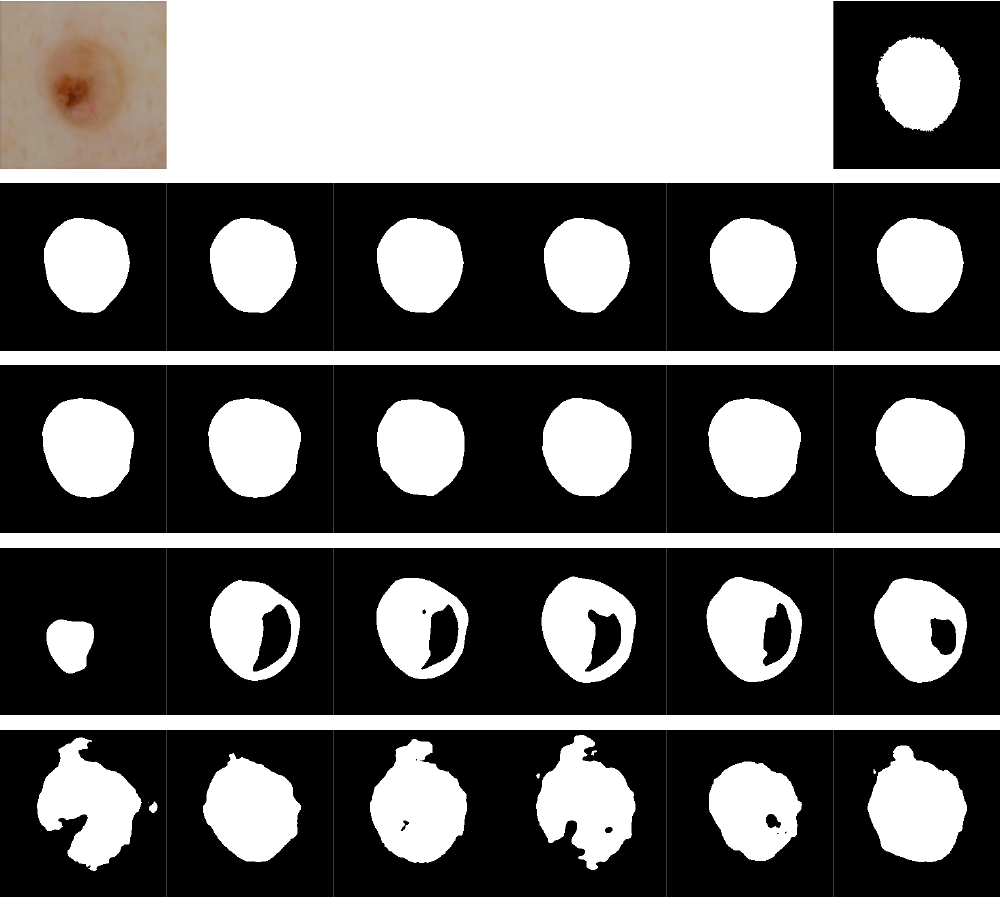}};
	\end{scope}
	
	\begin{scope}
	\node[anchor=south west, inner sep=0pt] (image1) at (0.6,0) {\includegraphics[width=0.4\textwidth]{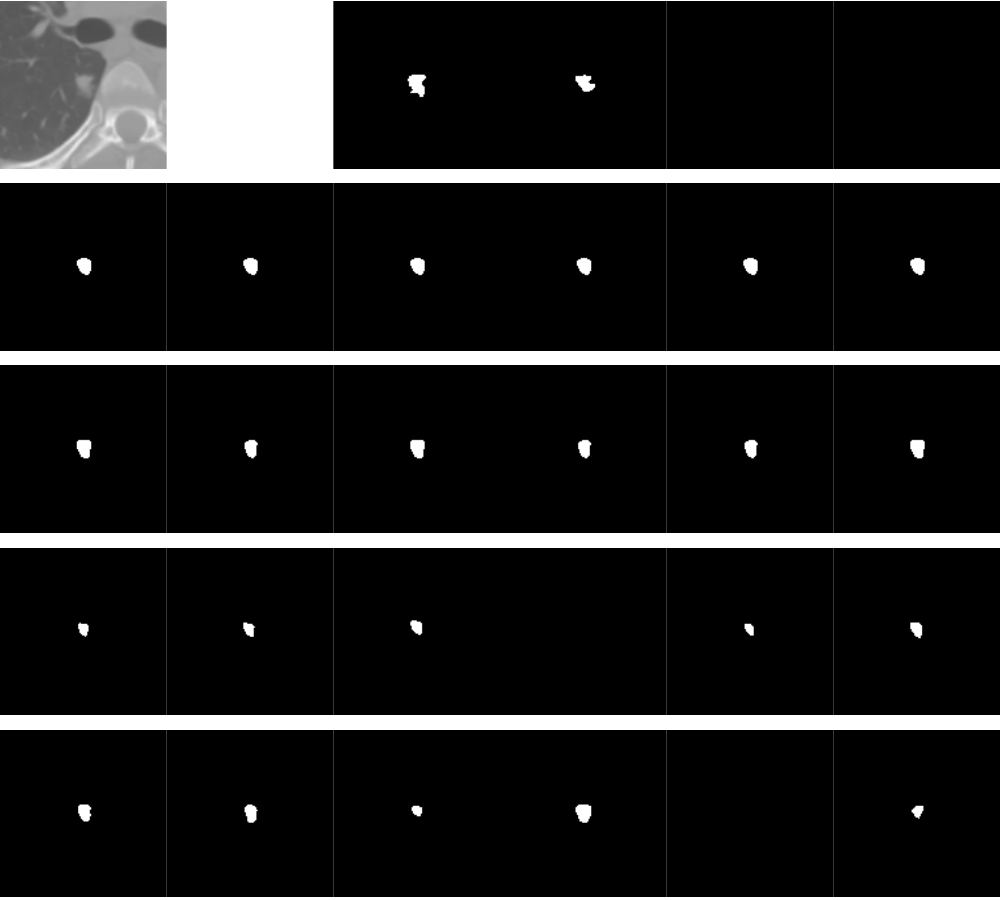}};
	\end{scope}
	
	\end{tikzpicture}
	\vspace{-2em}
	\caption{Samples from the probabilistic models. First row: Image and ground truth annotations from the skin dataset (left) and lung nodule dataset (right). Following rows: samples $\seg \sim \hat{p}(\y | \x, \X, \A)$ drawn from the various models. Sample diversity over the entire dataset shown next to the model name.}
	\label{fig:model_samples}
\end{figure}

\subsection{Sampling Segmentation Masks}
Fig.~\ref{fig:model_samples} shows segmentation masks $\y$ sampled from the trained models $\hat{p}(\seg|\x)$. The U-Net model is fully deterministic and does not offer any variation in samples. The sample diversity of the ensemble is limited by the number of constituent models (four in our experiment). The MC-Dropout and probabilistic U-Net allow fully random sampling and achieve a visually higher diversity. On the skin lesion dataset, where only one export annotation per image is available, models still produce diverse predictions. On the lung cancer dataset, samples from the MC-Dropout and probabilistic U-Net represent the annotator distribution well.

We measure the distance between the model distribution $\hat{p}(\y|\x, \X, \A)$ and the annotator distribution $p(\seg|\x)$ with the Generalized Energy Distance \cite{kohl2018probabilistic,monteiro2020stochastic,szekely2013energy}. The distance measure is calculated as
\begin{equation} \label{eq:ged}
D^2_{GED}(p, \hat{p}) = 2 \mathbb{E}_{y \sim p, \hat{y} \sim \hat{p}}\left[d(y,\hat{y})\right] 
- \mathbb{E}_{y, y' \sim p}\left[d(y,y')\right] 
- \mathbb{E}_{\hat{y},\hat{y}' \sim \hat{p}}\left[d(\hat{y},\hat{y}')\right] \enspace .
\end{equation}
We use $1 - \text{IoU}(\cdot, \cdot)$ as the distance $d$. A low $D^2_{GED}$ indicates similar distributions of segmentations. We approximate the metric by drawing up to 16 samples from both distributions, and sample with replacement. The results are shown in Fig.~\ref{fig:lidc_ged}. We observe that the annotator distribution is best approximated by the probabilistic U-Net, with MC-dropout and Ensemble closely behind; these pairwise ranks are significant $(\alpha = 0.01)$ with left-tailed t-tests. A deterministic U-Net architecture is not able to reproduce the output distribution. Our results are consistent with \cite{kohl2018probabilistic}, verifying our implementation. Following~\cite{monteiro2020stochastic}, we use the last term of \eqref{eq:ged} to assess the diversity of samples drawn from the model and note them in Fig.~\ref{fig:model_samples}. They reinforce the qualitative observations of sample diversity.

\subsection{Uncertainty estimates for active learning}
Instead of training the models with all available data $\{\mathbf X, \A\}$, we now start with a small random subset $\{\mathbf X_0, \A_0\}$. We train the model with this subset at iteration $t = 0$, and then add a set of $k$ images from $\{\mathbf X, \A\}$ to form $\{\mathbf X_{t+1}, \A_{t+1}\}$. Samples are selected based on the sum of pixel-wise entropies~\cite{lewis1994heterogeneous}. We repeat for $T$ iterations, benchmarking against a random sample selection strategy.


\begin{figure}[t]
    \centering
    \includegraphics[width=\linewidth]{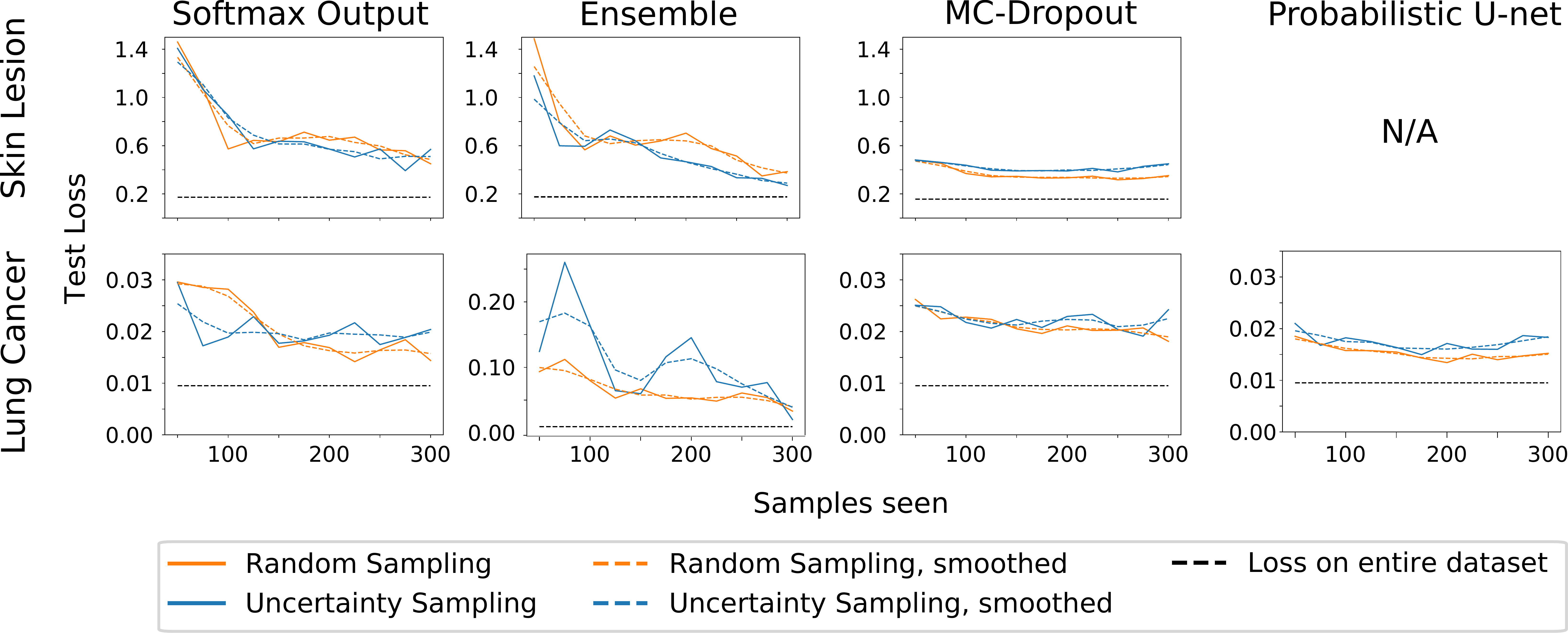}
    \caption{Learning curves for the four algorithms on both datasets. Note that the Probabilisitic U-net only applies to the ambiguous segmentation.}
    \label{fig:al_tablefig}
\end{figure}

For both skin lesion and lung cancer datasets, we start with a training size of 50 images, add  $k=25$ images at each iteration, and repeat $T=10$ times. The models are trained for 5000 gradient updates with a batch size of 16 and 32 for the respective datasets. Since annotations are costly and to speed up computations, no validation-loss based early stopping is used. The experimental setup has been picked to ensure meaningful model uncertainties for the data selection policy and to ensure convergence within each active learning iteration.

\begin{figure}[t]
	\centering
	\begin{tikzpicture}
	\node [anchor=mid, inner sep=0pt] (note) at (0.95,-0.4) {Image};
	\node [anchor=mid, inner sep=0pt] (note) at (3.25,-0.4) {Annotator 1};
	\node [anchor=mid, inner sep=0pt] (note) at (5.51,-0.4) {Annotator 2};
	\node [anchor=mid, inner sep=0pt] (note) at (7.78,-0.4) {Annotator 3};
	\node [anchor=mid, inner sep=0pt] (note) at (10.05,-0.4) {Annotator 4};
	
	\begin{scope}
	\node[anchor=south west, inner sep=0pt] (image1) at (0,0.0) {\includegraphics[width=0.9\textwidth]{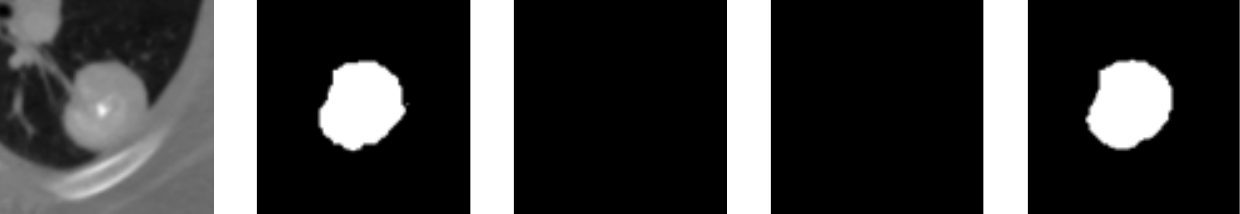}};
	\end{scope}
	
	\end{tikzpicture}
	\caption{An example of the ambiguous samples frequently selected for inclusion into the training set under the uncertainty-based data gathering strategy. This unseen sample was selected when 150 annotations were revealed. The group of expert annotators provided disagreeing segmentation masks, confirming the model uncertainty but providing little additional information to learn from.}
	\label{fig:disagree}
\end{figure}
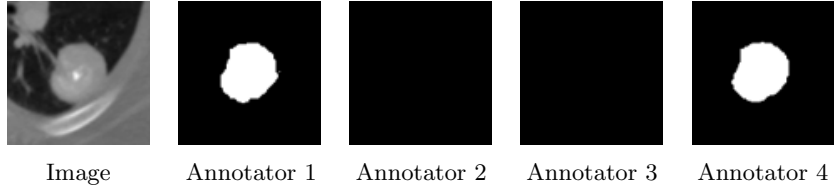

The learning curves in Fig.~\ref{fig:al_tablefig} show that random-based sampling leads to a faster reduction in test loss over the uncertainty-based sampling strategy for both datasets. We further investigated the samples selected by the uncertainty-based strategy by looking at the images which caused a large increase in the test error. One such image is shown in Fig.~\ref{fig:disagree}. 

\section{Discussion \& Conclusion}


Our results in Fig.~\ref{fig:uncertainty_by_condition} show that there is a clear relation between uncertainty estimates and segmentation error. The examples in Fig.~\ref{fig:uncertainty_samples} further highlight that areas of high uncertainty are not merely distributed around class boundaries, but also encompass areas with ambiguous labels. Fig.~\ref{fig:lidc_uncertainty_correl} shows that the uncertainty estimates obtained from the model are a good representation of the uncertainty of a group of expert annotators. We conclude that pixel-wise model uncertainty estimates give the practitioner a good indication of possible errors in the presented segmentation mask, allowing those predictions to be examined with care. 

The learning curves in Fig.~\ref{fig:al_tablefig} show that estimated uncertainty is not generally useful for selecting active learning samples, for any model or dataset. Our results depend on using the sum of pixel-wise entropies as a per-image entropy, which is correct for the softmax model, but only an approximation for the other models. This might impact our results. For the Lung Cancer dataset, all models estimate high uncertainty for the positive class, and the active learner thus selects images with a large foreground, skewing the proportion of classes represented in the training set. Furthermore, the selected images often have high annotator disagreement, illustrated in Fig.~\ref{fig:disagree}. If the active learner prefers sampling ambiguous images, it will be presented with inconsistent labels leading to harder learning conditions and poor generalisation. This may stem from an incorrect active learning assumption that annotations are noise-free and unambiguous, which is often not true. In conclusion, for a fixed budget of annotated images, we find no advantage in uncertainty-based active learning.


We observed similar behaviour of pixel-wise uncertainty estimates across all four segmentation models. The models differ in their ability to generate a distribution of distinct and coherent segmentation masks, with only the MC-dropout and probabilistic U-Net offering near unlimited diversity (see Fig.~\ref{fig:model_samples}). But these models are harder to implement, more resource-intensive to train, and require hyperparameter tuning.
The choice of model is ultimately application dependent, but our experiments show that even a simple U-net is competitive for the common task of assessing segmentation error. This agrees with~\cite{jungo2019assessing}, which compared uncertainty quantification models for unambiguous segmentation.

Our division of segmentation tasks into ambiguous and unambiguous considers it as "unambiguous" when a fundamentally ambiguous segmentation task is covered by a single annotator -- or potentially several annotators, but with only one annotator per image, as for the Skin Lesion dataset. Even if the underlying task \emph{is} ambiguous, the models considered in this paper inherently assume that it is \emph{not}, as there is no mechanism to detect annotator variance when every image is only annotated once. More fundamental modelling of segmentation ambiguity and uncertainty thus remains a highly relevant open problem.

To conclude -- is segmentation uncertainty useful? We find that uncertainty, even in the simplest models, reliably gives practitioners an indication of areas of an image that might be ambiguous, or wrongly segmented. Using uncertainty estimates to reduce the annotation load has proven challenging, with no significant advantage over a random strategy.

\subsubsection*{Acknowledgements.}

Our data was extracted from the “ISIC 2018: Skin Lesion Analysis Towards Melanoma Detection” grand challenge datasets~\cite{isic1,isic2}. The authors acknowledge the National Cancer Institute and the Foundation for the National Institutes of Health, and their critical role in the creation of the free publicly available LIDC/IDRI Database used here. This work was funded in part by the Novo Nordisk Foundation (grants no.~NNF20OC0062606 and NNF17OC0028360) and the Lundbeck Foundation (grant no.~R218-2016-883).

\appendix

\bibliographystyle{splncs04}
\bibliography{library}

\end{document}